%% file: arxiv.tex
\documentclass{article}

\usepackage[preprint]{neurips_2026}
\usepackage[utf8]{inputenc}
\usepackage[T1]{fontenc}
\usepackage{amsmath}
\usepackage{amssymb}
\usepackage{booktabs}
\usepackage{subcaption}
\usepackage{graphicx}
\usepackage{multirow}
\usepackage{xcolor}
\usepackage{tikz}
\usetikzlibrary{arrows.meta,fit,positioning}
\usepackage{hyperref}
\usepackage{listings}
\usepackage{needspace}
\usepackage{graphicx}
\usepackage{xparse}
\usepackage{enumitem}

\providecommand{\nolinenumbers}{}
\providecommand{\linenumbers}{}

\let\originalincludegraphics\includegraphics

\RenewDocumentCommand{\includegraphics}{O{} m}{%
  \IfFileExists{#2}{%
    \originalincludegraphics[#1]{#2}%
  }{%
    \fbox{\texttt{Missing figure: \detokenize{#2}}}%
  }%
}\lstnewenvironment{promptblock}
  {\nolinenumbers
   \lstset{
    basicstyle=\ttfamily\scriptsize,
    breaklines=true,
    breakatwhitespace=true,
    breakautoindent=false,
    breakindent=0pt,
    columns=fullflexible,
    keepspaces=true,
    showstringspaces=false,
    frame=single,
    framesep=4pt,
    aboveskip=0.4\baselineskip,
    belowskip=0.8\baselineskip,
    literate={~}{{\textasciitilde}}1 {—}{{\textemdash}}1
  }}
  {\linenumbers}

\newcommand{\promptheading}[1]{%
  \Needspace{8\baselineskip}%
  \medskip\noindent\textbf{#1}\par\nobreak
}

\title{LLMs as Acquisition Policies for Finite-Pool Materials Optimization:\
A Controlled Study}

\author{
  Dino--Rober Demir \\
  {\small\texttt{dino-rober.demir@polytechnique.org}}
  \And
  Florian Le Bronnec \\
  {\small\texttt{florian.lebronnec@riken.jp}}
  \And
  Rio Yokota \\
  {\small\texttt{rio.yokota@riken.jp}}
  \AND
  {\normalfont RIKEN Center for Computational Science, Tokyo, Japan}
}

\newcommand{\Xpool}{\mathcal{X}}
\newcommand{\Dt}{\mathcal{D}_t}
\newcommand{\Ct}{\mathcal{C}_t}

\newcommand{\argmax}{\operatorname*{arg\,max}}
\newcommand{\feconidataset}{\texttt{Fe--Co--Ni}}
\newcommand{\kerrdataset}{\texttt{Kerr Rotation}}
\newcommand{\coercivitydataset}{\texttt{Coercivity}}
\newcommand{\steelsdataset}{\texttt{Matbench Steels}}
\newcommand{\steelsdatasetid}{\texttt{matbench\_steels}}
\newcommand{\electrostraindataset}{\texttt{Electrostrain}}
\newcommand{\gemmamodel}{Gemma 4 31B}
\newcommand{\deepseekmodel}{DeepSeek-V4-Flash}
\newcommand{\qwenfamily}{Qwen3.5}
\newcommand{\qwenmid}{\qwenfamily-27B}
\newcommand{\qwenmidmoe}{\qwenfamily-35B-A3B}
\newcommand{\qwenbig}{\qwenfamily-397B-A17B}
\newcommand{\gemmamodelshort}{\textsc{Gemma4}}
\newcommand{\deepseekmodelshort}{\textsc{DS}}
\newcommand{\qwenmidshort}{\textsc{Qw27B}}
\newcommand{\qwenmidmoeshort}{\textsc{Qw35B}}
\newcommand{\qwenbigshort}{\textsc{Qw397B}}

\newcommand{\meanstd}[2]{#1\,{\scriptsize$\pm$\,#2}}

\begin{document}

\maketitle

\begin{abstract}
  Discovering materials with desirable properties often requires searching large candidate spaces while experimental or computational evaluations remain costly. Active learning addresses this challenge by using previous observations to select which candidate to evaluate next, typically through probabilistic surrogate models. We investigate whether open-weight large language models (LLMs) can serve as standalone acquisition policies in this setting. We evaluate five LLMs across four retrospective finite-pool materials optimization tasks under different candidate-presentation strategies and compare them with random selection and conventional Gaussian-process methods. LLM policies generally reach the global optimum in fewer iterations than random selection, indicating that they provide a useful acquisition signal without task-specific training. Their performance relative to Gaussian-process methods is mixed: conventional acquisition performs better on most tasks, while LLMs match or outperform it in some settings. Performance varies substantially across tasks, models, initializations, and candidate presentations, with no LLM approach performing best across all tasks. Overall, open-weight LLMs show potential as acquisition policies for finite-pool materials search, although their reliability remains sensitive to the task and to how candidates and scientific context are presented.
\end{abstract}

\section{Introduction}

Materials science has long been a thriving field of research, and the discovery of new materials is a key driver of technological progress \citep{depablo2019new,jain2013materialsproject}. However, conventional experimental and computational screening remains costly and time-consuming \citep{schmidt2019recent,pyzerknapp2022accelerating}. In the last decade, machine learning has emerged as a promising way to address these challenges by predicting materials properties and guiding the search for new materials \citep{butler2018machine,ramprasad2017machine}. In this work, we evaluate whether pretrained LLMs can translate their learned priors into useful acquisition policies for AI-guided materials discovery and study their reliability relative to established materials active-learning methods.

Active learning and Bayesian optimization are standard tools for materials discovery when experimental or computational labels are expensive \citep{lookman2019active}. In a typical surrogate-based closed-loop experiment, an optimizer observes a small set of labeled candidates, fits a predictive model, and selects the next candidates to evaluate by maximizing an acquisition function that captures the trade-off between exploration and exploitation over the candidate pool. Gaussian processes (GPs), together with acquisition functions such as expected improvement ($\mathrm{EI}$) or upper confidence bound ($\operatorname{UCB}$), remain strong baselines in Bayesian optimization because they explicitly model both predicted performance and epistemic uncertainty \citep{kushner1964new,srinivas2010gaussian,pml1Book}.

Large language models (LLMs) offer a different route to candidate selection. Recent reasoning-focused LLMs have demonstrated strong performance on scientific reasoning benchmarks \citep{openai2024o1,guo2025deepseekr1}. The extent of their knowledge, their reasoning capabilities, and their applicability to materials science are still being explored. These capabilities make LLMs attractive for sequential candidate selection, but their reliability across materials problems remains unclear.

We conduct this evaluation on controlled retrospective materials active-learning benchmarks, where LLMs are prompted with the observed history, a candidate list, and a natural-language description of the scientific objective. At each iteration, the LLM policy selects exactly one unobserved composition from a finite candidate pool, receives the corresponding property value from the dataset, and repeats until it selects the globally optimal candidate. This protocol allows direct comparison between classical GP acquisition functions, standalone LLM selection, and LLM acquisition variants under identical experimental conditions.

We organize the study around three questions:
\begin{itemize}[leftmargin=1.0em]
  \item How competitive and reliable are LLM acquisition policies compared with GP-EI?
  \item How do candidate presentation and materials-specific semantic context affect LLM selection?
  \item How do LLM-selected candidates compare to those selected by GP-based Bayesian optimization?
\end{itemize}

We address these questions by evaluating five open-weight LLMs across four materials optimization tasks and a range of experimental conditions. Across these experiments, we observe a useful but variable acquisition signal and examine how its reliability is shaped by the task, candidate presentation, and semantic context.

\section{Related Work}

\paragraph{Active learning for materials discovery.} Materials discovery remains costly, motivating machine-learning approaches to predict properties and guide candidate selection \citep{depablo2019new,schmidt2019recent,butler2018machine}. Retrospective materials benchmarks provide a controlled setting for evaluating active-learning methods by treating a labeled dataset as an oracle \citep{wang2022benchmarking}. While they do not fully reproduce experimental deployment, they allow repeated trials from different initializations and direct comparison between acquisition policies. This makes them particularly well suited to our goal: isolating how LLM-based candidate selection behaves relative to established acquisition strategies, without run-to-run experimental variation or changes to the candidate pool.

\paragraph{LLMs for active learning and optimization.} LLMs have been incorporated into active-learning and optimization pipelines in several roles. In machine learning, LLMs have been used to select examples for training downstream models \citep{bayer2026activellm,parkar2024selectllm} and to prune unlabeled pools before standard active-learning acquisition \citep{azeemi2026activeprune}; \citet{xia2025selection} review this broader literature. In black-box Bayesian optimization, \citet{chang2025llinbo} introduce LLINBO, a hybrid framework that combines LLM-suggested query points with a statistical surrogate. These studies illustrate how LLMs can support data acquisition, either by identifying useful training examples or by complementing conventional optimizers. However, they primarily address language tasks or general black-box optimization rather than materials discovery specifically.

\paragraph{LLMs for materials optimization.} Prior work in chemistry and materials optimization has coupled LLMs with probabilistic surrogates \citep{kristiadi2024sober,rankovic2025gollum} and used LLM-generated hypotheses or tool-enabled agents to explore large implicit design spaces, with or without Bayesian optimization \citep{bora,cisse2026closedloop}. Unlike these methods, we focus on standalone LLM acquisition from explicit finite candidate pools. The most closely related study by \citet{wang2026trainingfree} uses a generation-and-matching approach (see Section~\ref{sec:generation-and-matching}) within small candidate pools containing 73 to 323 points with closed-weight LLMs. We evaluate this protocol on open-weight LLMs and compare it with other LLM policies.

\section{Background}

\paragraph{Pool-based active-learning protocol.}

Let $\Xpool = \{x_i\}_{i=1}^N$ be a finite pool of candidate materials with known labels $\{y_i\}_{i=1}^N$, hidden from the policy until queried. For each dataset and acquisition-policy configuration, candidate $x_i \in \mathbb{R}^d$ denotes the representation provided to the policy, where $d$ depends on the dataset and input representation, and $y_i \in \mathbb{R}$ is its scalar label. The goal is to maximize the target value $y$ among $\{y_i\}_{i=1}^N$. At iteration $t$, the policy observes
\begin{equation}
  \Dt = \{(x_j, y_j): j \in \mathcal{I}_t\},
\end{equation}
where $\mathcal{I}_t$ is the set of previously queried indices, and selects one unobserved candidate
\begin{equation}
  i_t \in \{1,\ldots,N\}\setminus\mathcal{I}_t.
\end{equation}
The oracle returns $y_{i_t}$, and the observed set is updated to $\mathcal{D}_{t+1}=\Dt\cup\{(x_{i_t},y_{i_t})\}$ (and $\mathcal{I}_{t+1}=\mathcal{I}_t\cup\{i_t\}$). On a fixed seed, every method is initialized with the same single initial point. Moreover, each subsequent acquisition adds exactly one newly queried index to \(\mathcal{I}_t\) and the corresponding candidate-label pair to \(\mathcal{D}_t\).

\paragraph{Gaussian-process baseline.}

We compare against GP-based Bayesian optimization baselines. At each iteration, the GP is fit on the observed data $\Dt$, producing a posterior mean $\mu_t(x)$ and posterior standard deviation $\sigma_t(x)$ for each unobserved candidate. We use the Expected Improvement ($\mathrm{EI}$) acquisition function \citep{jones1998efficient}. It selects:

\begin{equation}
  i_t^{\mathrm{EI}} \in \argmax_{i \in \{1,\ldots,N\} \setminus \mathcal{I}_t} \mathrm{EI}_t(x_i),
\end{equation}
where
\begin{equation}
  \mathrm{EI}_t(x) = (\mu_t(x)-b_t)\Phi(z_t(x)) + \sigma_t(x)\phi(z_t(x)), \quad
  z_t(x) = \frac{\mu_t(x)-b_t}{\sigma_t(x)}, \quad
  b_t=\max_{j\in\mathcal{I}_t}y_j,
\end{equation}
When $\sigma_t(x) = 0$, we set $\mathrm{EI}_t(x)=(\mu_t(x)-b_t)_+$. Here $\Phi$ and $\phi$ denote the standard normal CDF and PDF. $\mathrm{EI}$ combines predicted improvement over the incumbent with posterior uncertainty. In the canonical form used here, it does not require an explicit exploration coefficient such as the $\beta$ parameter in GP-UCB.

\paragraph{Random baseline.}

We also compare our acquisition policies to a random baseline, in which the points are sampled randomly among the remaining unlabeled data:
\begin{equation}
  i_t^{\mathrm{Random}} \sim \mathcal{U}({\{1,\ldots,N\} \setminus \mathcal{I}_t}).
\end{equation}

\section{LLM-Based Candidate Selection}

\paragraph{LLM acquisition policies.}
An LLM acquisition policy \(\pi_{\mathrm{LLM}}\) receives the observed history, a candidate list, and task instructions, then returns the index of one candidate to evaluate next. Unlike GP-EI, which assigns an explicit acquisition score to every remaining candidate after fitting the surrogate, the LLM policies considered here make direct, potentially list-dependent choices and do not construct a pool-wide acquisition-score vector. Obtaining a separate LLM assessment for every candidate would require additional inference proportional to the pool size and is not necessary here. It selects:
\begin{equation}
  i_t^{\mathrm{LLM}}
  \sim \pi_{\mathrm{LLM}}
  \left(
  \cdot \mid \Dt, \Ct, p
  \right),
\end{equation}

where $\Ct$ is an ordered presentation to the LLM of unlabeled data, \(i_t^{\mathrm{LLM}}\) denotes its final selected candidate index rather than the output of an individual LLM call, and $p$ is the prompt and decoding configuration. Unless otherwise specified, we use materials-aware prompts: candidates are represented using semantically meaningful feature labels, such as element names and target-property names, together with a materials-specific description of the task. These system prompts depend on the dataset and optimization objective; examples are provided in Appendix~\ref{app:prompts}.

\paragraph{Candidate-selection protocols.}
Each candidate-selection protocol induces an LLM-based acquisition policy over the remaining pool. Our primary protocol is whole-pool selection: it presents all remaining candidates to the model in a single, randomly ordered list and asks the model to return the displayed index of the candidate it considers most promising. A protocol may instead involve multiple nested LLM calls, but it always yields a single candidate index for evaluation. We introduce the batch-tournament protocol alongside its analysis in Section~\ref{sec:candidate-protocol}; implementation details for both protocols are provided in Appendix~\ref{app:implementation-details}.

\section{Experimental setup}

\paragraph{Datasets.}

We evaluate our methods on four optimization tasks spanning three materials datasets. Each task is defined by a finite candidate pool in which every candidate has a vector-valued representation and a scalar objective to maximize.

\paragraph{\feconidataset.} The \feconidataset{} combinatorial thin-film benchmark introduced by \citet{wang2022benchmarking} contains 921 experimentally measured ternary compositions, i.e., $\mathrm{Fe}_x\mathrm{Co}_y\mathrm{Ni}_{1-x-y}$. Although the original library also includes X-ray diffraction data and measured magnetic properties, we use composition-only inputs. We consider two target properties: Kerr rotation in $\mathrm{mrad}$ and magnetic coercivity in $\mathrm{mT}$. They define two separate optimization tasks called \kerrdataset{} and \coercivitydataset{}, respectively. The original benchmark describes coercivity as the more challenging target because of its more complex objective landscape.

\paragraph{\steelsdataset.} Introduced as \steelsdatasetid{} in the Matbench v0.1 benchmark by \citet{dunn2020benchmarking}, this dataset is defined as a supervised regression task, with yield strength in $\mathrm{MPa}$ as the target property. It contains 312 composition-only entries with 14 features. We follow the retrospective pool-based optimization framing also used for \steelsdataset{} by \citet{wang2026trainingfree}, in which we treat the labeled dataset as a finite oracle and ask each acquisition policy to maximize yield strength.

\paragraph{\electrostraindataset.}
This dataset is derived from the BaTiO$_3$-based piezoelectric active-learning study of \citet{yuan2018accelerated}, also reviewed by \citet{lookman2019active}. It combines the 61 initial measurements with the 20 compositions synthesized over five active-learning iterations, yielding a finite pool of 81 candidates. The original prospective search considered solid solutions of the form $(\mathrm{Ba}_{1-x-y}\mathrm{Ca}_x\mathrm{Sr}_y)(\mathrm{Ti}_{1-u-v}\mathrm{Zr}_u\mathrm{Sn}_v)\mathrm{O}_3$; the initial measurements additionally contain a small number of Cd-containing compositions. Each candidate is represented by eight cation-fraction columns (Ba, Ca, Sr, Cd, Ti, Zr, Sn, and Hf) and seven auxiliary physicochemical descriptors, for 15 input features in total; Hf is identically zero in the observed pool. The objective is to maximize the bipolar electrostrain measured at $20\, \mathrm{kV\,cm^{-1}}$. Further details are provided in Appendix~\ref{app:electrostrain}.

\input{tables/datasets}

\paragraph{Models.}

To facilitate reproducibility and limit inference costs, we restrict our experiments to five open-weight models spanning three model families and several scales. This diversity allows us to assess the sensitivity of our findings to model family and scale. These include \gemmamodel{}, \deepseekmodel{} (284B-A13B), \qwenmid{}, \qwenmidmoe{}, and \qwenbig{} \footnote{Throughout the whole paper, \(^{*}\) and \(^{\dagger}\) near \qwenbig{} runs results denote proxy runs due to inference costs; see Appendix~\ref{app:model-settings}} \citep{gemma4, deepseekv4, qwen35blog}. Further model and inference details are provided in Appendix~\ref{app:model-settings}.

\section{Whole-Pool LLM Acquisition}
\label{sec:standalone}

We first evaluate whole-pool LLM selection against random selection and GP-EI. Table~\ref{tab:main-results} additionally reports batch-tournament results for completeness; these are analyzed in the next section.

\begin{figure}[t]
  \centering
  \includegraphics[width=\linewidth]{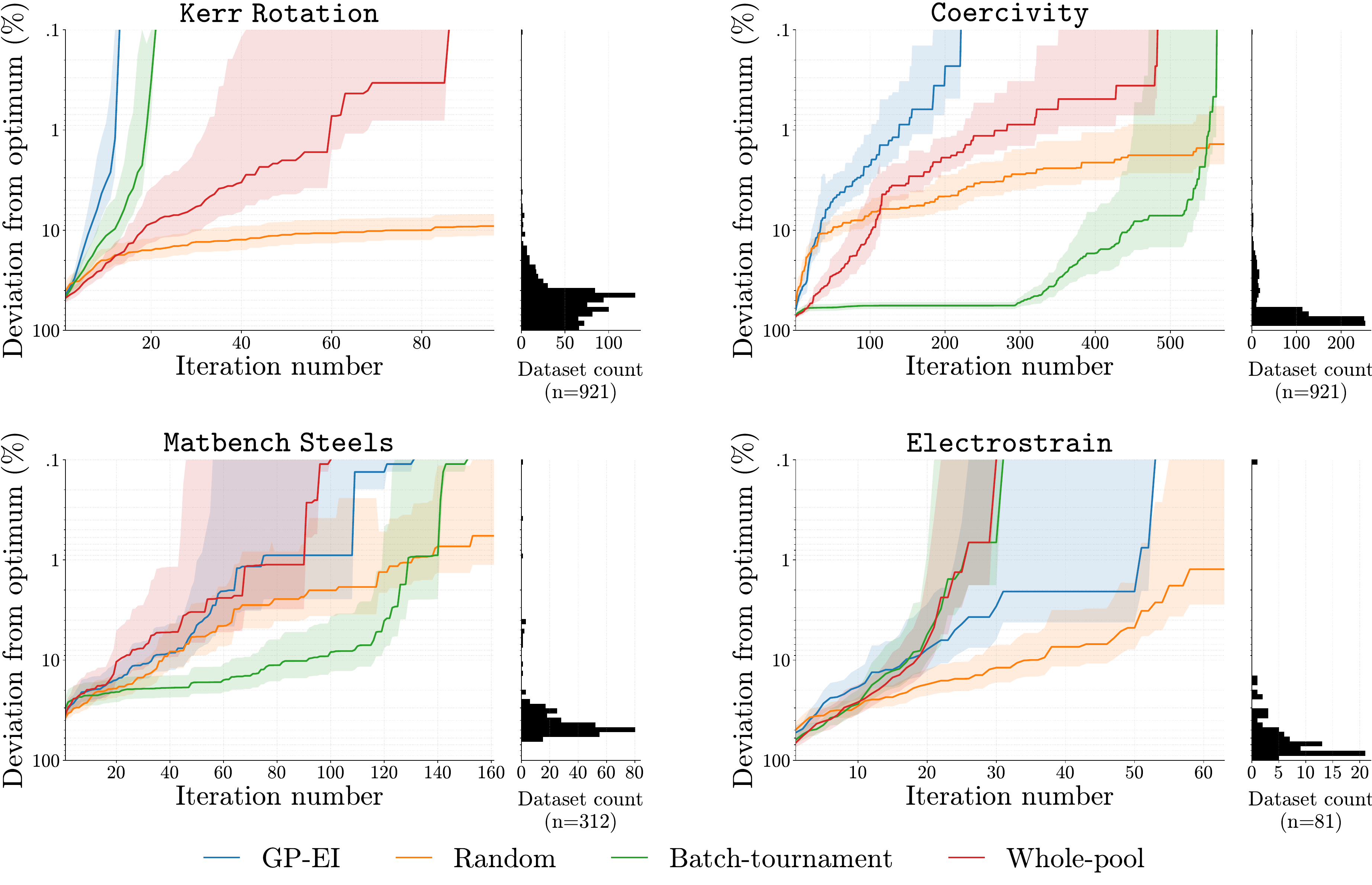}
  \caption{Best-so-far optimization trajectories using \gemmamodel{} for the LLM policies. At iteration $t$, each curve shows the mean percentage gap $100(y^\star-b_t)/|y^\star|$ across the 25 shared initialization seeds. Shading denotes pointwise normal-approximation 95\% confidence intervals, computed as $\bar{g}_t \pm 1.96s_t/\sqrt{25}$. We show \gemmamodel{} because it ranks among the strongest LLMs across all four tasks.}
  \label{fig:main-performance}
\end{figure}

\input{tables/main-results}

\paragraph{Performance and variability.}

Table~\ref{tab:main-results} compares whole-pool LLM selection with GP-EI and random selection. GP-EI achieves the lowest mean iterations to optimum on three of the four tasks: \kerrdataset{}, \coercivitydataset{}, and \steelsdataset{}. \electrostraindataset{} is the exception, with three of the five LLMs requiring fewer iterations on average than GP-EI. Every whole-pool LLM also achieves a lower mean than random selection on every task. Whole-pool selection therefore provides a useful acquisition signal, although it does not consistently match GP-EI across tasks. Variability also tends to follow similar trends to those of mean iterations to optimum, although the magnitude of the deviation depends strongly on the model and task.

\paragraph{Consistency.}

Whole-pool model rankings are not consistent across tasks. \deepseekmodel{} achieves the lowest mean iterations to optimum on \electrostraindataset{}, but the highest on \kerrdataset{} and \steelsdataset{}. By contrast, \gemmamodel{} remains near the top across all four tasks. Neither model scale nor architecture provides a clear ordering even within the \qwenfamily{} family: \qwenmid{} consistently outperforms \qwenmidmoe{}, while the available results for \qwenbig{} show no systematic advantage over \qwenmid{}. Therefore, neither single-task performance nor nominal model scale yields a consistent cross-task ranking. These reversals motivate examining whether candidate presentation, rather than model choice alone, changes acquisition performance.

\section{Batch-Tournament LLM Acquisition}
\label{sec:candidate-protocol}

\paragraph{Batch-tournament LLM selection.}
LLMs are known to struggle with long contexts \citep{liu2024lost}. To control the number of candidates shown in each call, batch-tournament selection randomly splits the unobserved pool into batches of size $b$. The LLM selects one winner from each batch, after which the winners are rebatched and the process repeats until the remaining set is small enough for a final LLM selection (see Figure~\ref{fig:batch-tournament-schematic}). This protocol gives every candidate an opportunity to enter the tournament while limiting the size of each individual comparison. When the batch size equals the pool size, it reduces to whole-pool selection. Unless otherwise specified, we use (b=20) across all experiments, so the LLM receives at most 20 candidates per call.

\begin{figure}[b]
  \centering
  \includegraphics[width=\linewidth]{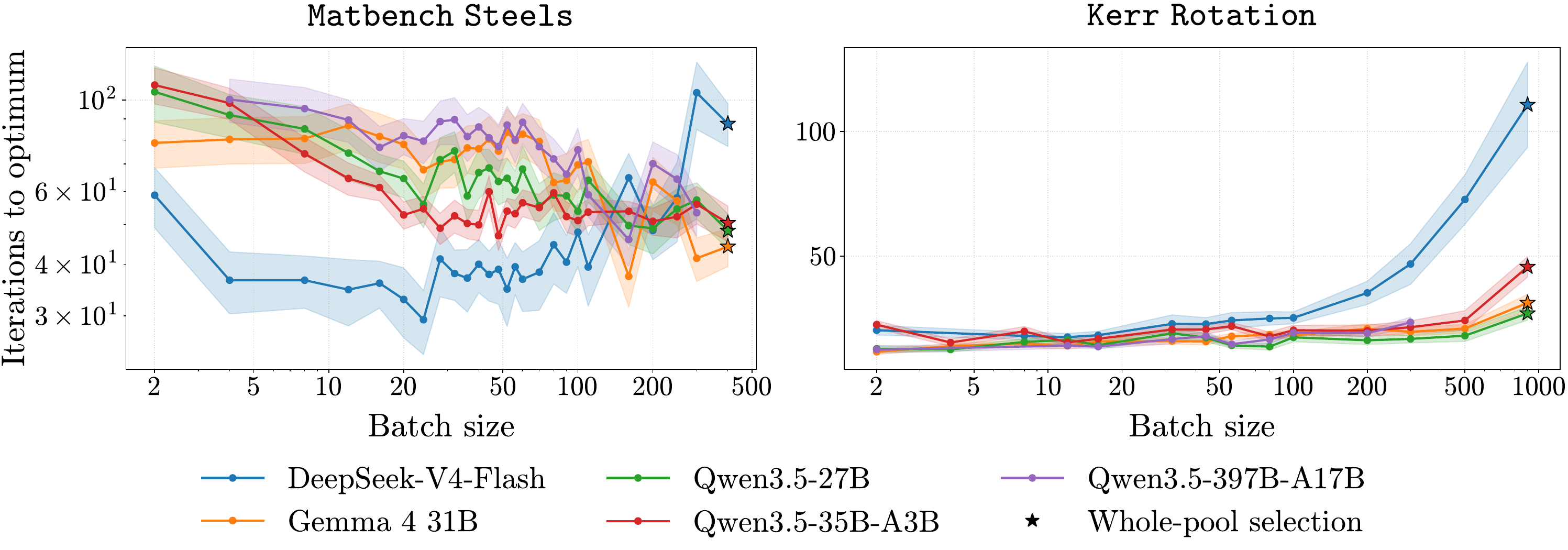}
  \caption{Batch-size sensitivity of tournament selection on \steelsdataset{} and \kerrdataset{} for runs of 25 seeds. Curves report the mean number of iterations to the optimum across seeds, with uncertainty shown shaded as the mean $\pm$ $\mathrm{std}/\sqrt{n_{\mathrm{seeds}}}$.}
  \label{fig:batch-size}
\end{figure}

\paragraph{Small batch-tournament selection behaves differently from whole-pool selection.}

Table~\ref{tab:main-results} shows that, on \kerrdataset{} and \electrostraindataset{}, batch-tournament selection achieves a lower mean iterations to optimum and a lower standard deviation across almost every model. However, the same cannot be said on \steelsdataset{} and \coercivitydataset{}, which are the two tasks that generally take more iterations to converge. Overall, comparable mean iterations to optimum tends to yield higher variability for the whole-pool selection compared to a small batch-sized tournament, which is closer to that of GP-EI. Regardless, neither selection policy uniformly dominates the other across the four tasks. The trajectories in Figure~\ref{fig:main-performance} show that even in early stages, sometimes batch-tournament is one of the best policies and sometimes it is even worse than random selection for \gemmamodel{}.

\paragraph{Batch-size effects are task-dependent, with model-specific deviations.} Figure~\ref{fig:batch-size} shows that no batch size performs consistently well across tasks. For the \qwenfamily{} models and \gemmamodel{}, increasing the batch size generally reduces the mean iterations to optimum on \steelsdataset{}, but increases it on \kerrdataset{}. Smaller batches therefore tend to work better on the latter task, whereas the first one benefits from a wider view of the candidate pool (\coercivitydataset{} follows a trend similar to that of \steelsdataset{}, and \electrostraindataset{} similar to that of \kerrdataset{}, see Figure~\ref{fig:batch-size-coer}). The pattern nevertheless also depends on the model: \deepseekmodel{} follows the broad trends at small and intermediate batch sizes, but deteriorates sharply at the largest sizes. This deviation may be related to its stronger position bias, examined in Section~\ref{sec:position-bias}.

\paragraph{Additional candidate-selection protocols performance.}
Similar experiments were conducted with a generation-and-matching protocol. The results are reported in Appendix~\ref{sec:generation-and-matching}.

\section{Diagnostics of LLM Acquisition Behavior}

\subsection{Position bias}
\label{sec:position-bias}

LLM performance on long-context information retrieval and question-answering tasks has been shown to depend strongly on the position of relevant information, often degrading when that information occurs in the middle of the context \citep{liu2024lost}. We therefore examine where in the batch each selected candidate occurs and quantify position bias as the $\mathrm{KL}$-divergence between the empirical distribution and a uniform-selection baseline, with full details provided in Appendix~\ref{app:position-bias}.

\begin{figure*}[t]
  \centering
  \begin{subfigure}[c]{0.70\textwidth}
    \centering
    \includegraphics[width=\linewidth]{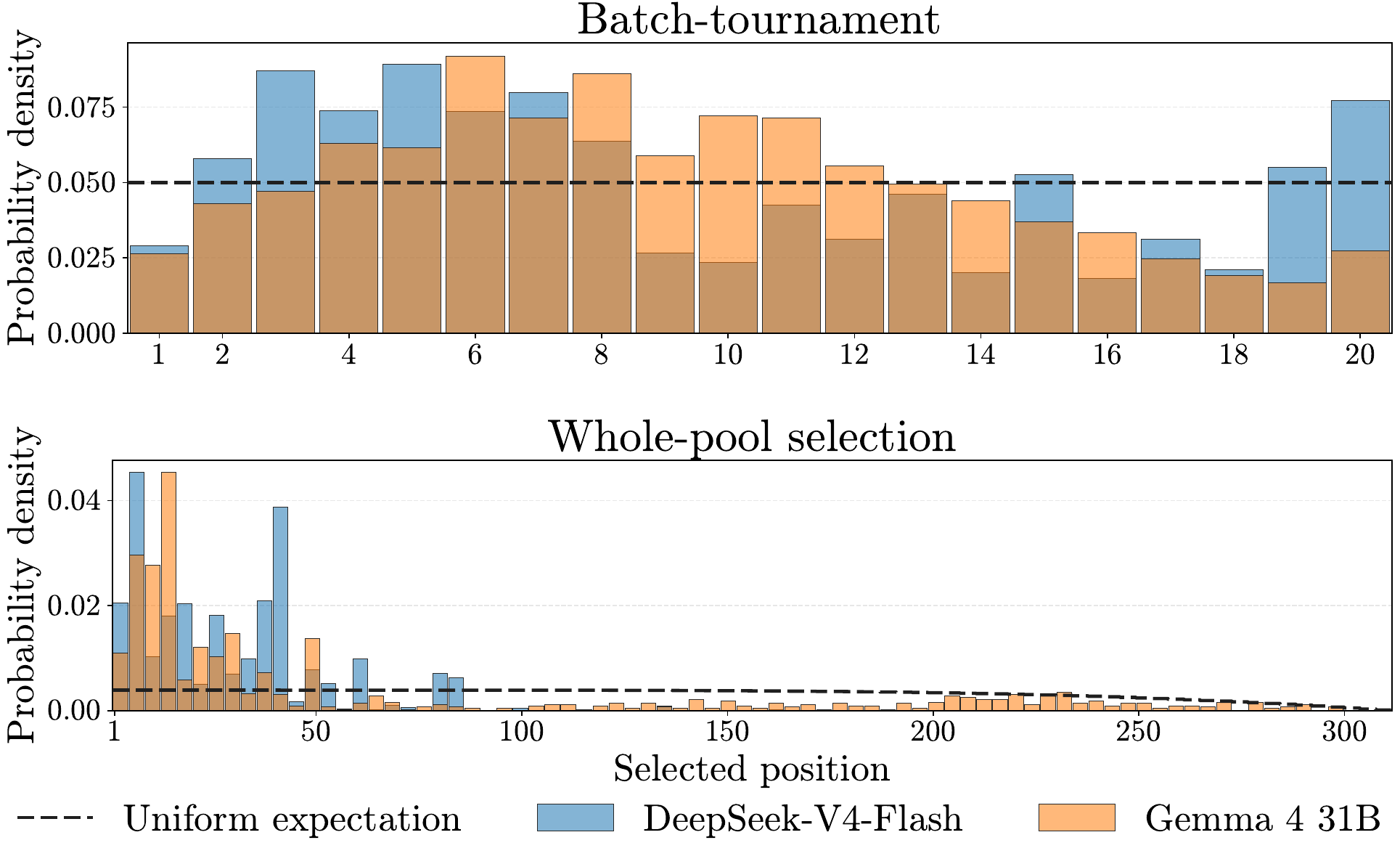}
    \caption{Position distributions under batch-tournament and whole-pool selection.}
    \label{fig:position-bias-distributions}
  \end{subfigure}\hfill
  \begin{subfigure}[c]{0.27\textwidth}
    \centering
    \input{tables/position-bias-steels}
    \caption{Whole-pool position bias on \steelsdataset{}.}
    \label{fig:position-bias-metrics}
  \end{subfigure}
  \caption{Position bias for candidates selected by \deepseekmodel{} and \gemmamodel{} on \steelsdataset{}. Dashed lines in panel~(a) indicate the uniform-selection expectation, which may not be horizontal because the pool shrinks across iterations. Panel~(b) reports $D_{\mathrm{KL}}(q\,\|\,u)$ for all evaluated models.}
  \label{fig:position-bias}
\end{figure*}

\paragraph{Position bias depends on the model and the setting.}

Figure~\ref{fig:position-bias} shows a non-uniform positional selection pattern under whole-pool prompting on \steelsdataset{} that occurs to a much lesser extent under acquisition within small batches. The effect is particularly pronounced for \deepseekmodel{}, whose selections concentrate near the beginning of the candidate list, with almost no selections beyond the first 100 positions. \gemmamodel{} selects from a broader portion of the list but still exhibits localized peaks. The complete $\mathrm{KL}$-divergence results reported in the appendix Table~\ref{tab:position-bias-metrics-full} show that the magnitude of this non-uniformity varies substantially across models and tasks. Candidate ordering should therefore be considered a potential source of sensitivity when dealing with large contexts.

\subsection{Materials Context}
\label{sec:materials-context}

\begin{figure}[h]
  \centering
  \includegraphics[width=\linewidth]{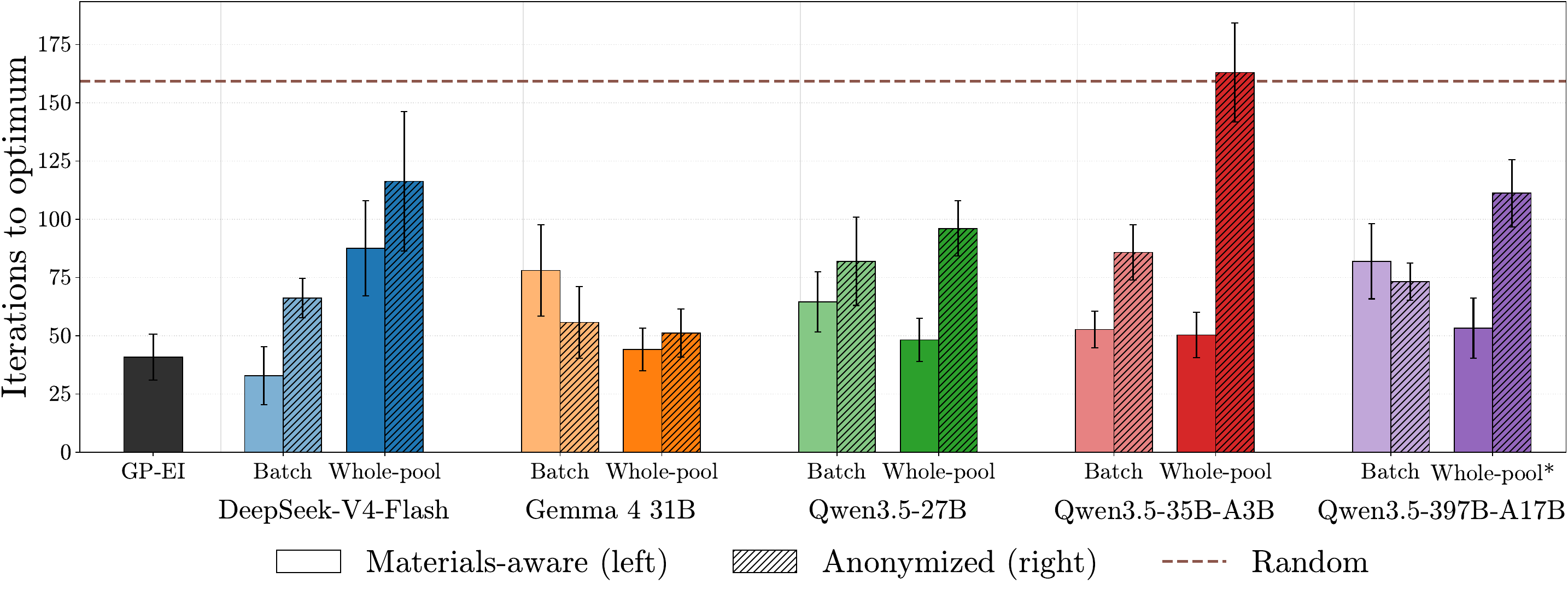}
  \caption{Optimization performance on \steelsdataset{}, with and without materials context across 25 initializations. Bars report the mean iteration at which the optimum is reached, with normal-approximation 95\% confidence intervals, computed as mean $\pm$ $1.96 \,\mathrm{std}/\sqrt{n_{\mathrm{seeds}}}$. We also report the mean iterations to optimum of random selection with a dotted line.}
  \label{fig:context-comparison}
\end{figure}

Prior work has suggested that LLMs can iteratively propose solutions from previously evaluated solution-value pairs and adapt their in-context priors in regression and bandit settings \citep{yang2024large,codaforno2023meta}. In materials science, LLMs may encode qualitative knowledge about composition-property relationships, synthesis constraints, or known alloying trends. We test whether this semantic context affects acquisition performance on \steelsdataset{} by comparing materials-aware and anonymized-feature prompts under both whole-pool and batch-tournament selection, using the same 25 initializations shown in Figure~\ref{fig:context-comparison}.

\paragraph{Prompt conditions.}
In the materials-aware condition used throughout the main experiments, candidates retain semantically meaningful feature labels and the prompt describes the materials optimization task. In the anonymized condition, the same numerical features are denoted $x_1, \ldots, x_d$, and the task is described only as maximizing an unknown scalar target. The anonymized prompt does not mention materials, alloys, elements, chemistry, or physical property names. Thus, the conditions differ in their semantic context while preserving the numerical candidate information. Prompt examples are provided in Appendix~\ref{app:prompts}.

\paragraph{The effect of materials-aware prompting depends on the candidate-selection protocol.}

Figure~\ref{fig:context-comparison} shows that LLMs retain meaningful optimization ability without explicit materials context: nine of the ten anonymized model--protocol configurations reach the optimum in fewer mean iterations than random selection. Materials-aware prompting nevertheless yields a lower mean in eight of the ten comparisons, including all five whole-pool settings. Under batch-tournament selection, anonymized prompting yields lower means for \gemmamodel{} and \qwenbig{}, although the confidence intervals of the two prompt conditions overlap in both cases. These results indicate that numerical features and observed outcomes alone provide a useful signal for black-box optimization, while semantic descriptions of the features and scientific objective provide additional value in many materials active-learning settings. This pattern is consistent with pretrained LLMs mobilizing materials-relevant prior knowledge when semantic context is available.

\subsection{GP-Relative Acquisition Diagnostic}
\label{sec:gpdiagnostics}

Examining iterations to optimum performance does not reveal what kind of acquisition policy a LLM implements. To characterize selection behavior, we investigate whether LLM selections can be compared to those of GP methods by visualizing their behavior in a two-dimensional PCA space.

\begin{figure}[t]
  \centering
  \includegraphics[width=0.9\linewidth]{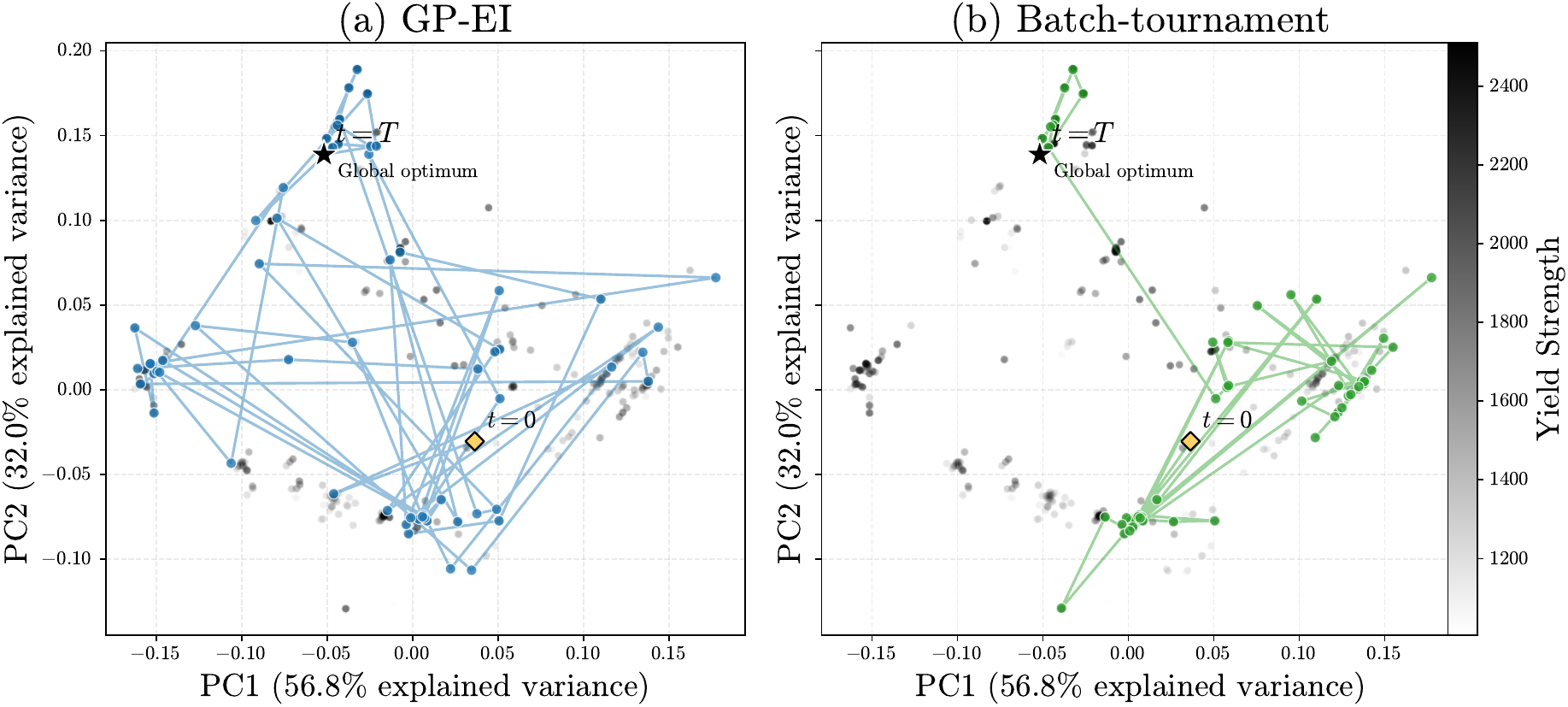}
  \caption{Explorative behavior of two selection protocols on \steelsdataset{} for seed 4 with \gemmamodel. Dots represent points of the pool projected onto the first two principal components, with the PCA being scaled over every pool entry. On this seed, GP-EI and batch-tournament reach the optimum after 60 and 56 iterations, respectively.}
  \label{fig:exploration-exploitation}
\end{figure}

\paragraph{Comparison with GP-like exploration-exploitation behavior.} Figure~\ref{fig:exploration-exploitation} shows a single illustrative trajectory on \steelsdataset{}. In this run, GP-EI initially samples across the composition space before concentrating its selections, whereas \gemmamodel{} focuses earlier on a narrower region while still occasionally selecting distant candidates. This example suggests a more irregular transition between exploration and exploitation for the LLM policy. Even though the LLM selections are informed enough to outperform random selection, they do not resemble a regular GP-EI policy. Further analysis in Appendix~\ref{sec:ucb-fit} provides additional results consistent with this observation.

\section{Conclusion}

Across four retrospective finite-pool materials benchmarks, our results show that pretrained open-weight LLMs provide a genuine acquisition signal. LLM policies repeatedly outperform random selection, including when candidate features and objectives are anonymized. On \steelsdataset{}, adding materials-specific feature names and task descriptions improves mean performance in most evaluated settings and in every whole-pool comparison. Together, these results provide evidence that pretrained LLMs carry materials-relevant priors that can inform sequential candidate selection.

These priors do not yet produce consistently competitive acquisition policies. GP-EI remains stronger on most tasks under whole-pool selection, while LLM rankings vary across tasks and do not follow predictably from model size or architecture. Performance is also sensitive to the candidate-selection protocol, batch size, initialization, and candidate ordering, with some models exhibiting substantial position bias. The usefulness of the LLM acquisition signal therefore depends strongly on how the candidate pool and scientific context are presented.

Within our focused scope of retrospective finite-pool optimization, these findings establish standalone LLM acquisition as a credible direction for AI-guided materials discovery. The controlled comparisons developed here provide an initial empirical basis for studying how pretrained scientific priors can be translated into more reliable acquisition decisions across tasks and at larger scales.
\clearpage

\bibliographystyle{plainnat}
\bibliography{references}
\clearpage
\appendix

\section{Limitations}
\label{sec:limitations}

\paragraph{Realistic pool sizes.}

Our experiments use retrospective benchmarks with candidate pools of at most 921 points, whereas real materials-discovery campaigns may involve hundreds of thousands or millions of candidates. For example, BORA's hydrogen-evolution benchmark initially contains 98,423,325 feasible experimental combinations \citep{bora,burger2020mobile}, while the original \electrostraindataset{} study considered around 605,000 compositions \citep{yuan2018accelerated}. Our conclusions should therefore be interpreted in light of these substantial differences in scale.

\paragraph{Candidate selection at scale.}

Scale changes the relative appeal of the candidate-selection protocols. In a very large implicit space such as BORA's, or in a continuous space, generating a promising design is natural because presenting every possible candidate is impossible. For a large explicit pool, generation-and-matching may likewise become more useful because a denser pool is more likely to contain a close match to an LLM-generated proposal, provided that matching can be performed efficiently. On the small pools considered by \citet{wang2026trainingfree} of 73 to 323 candidates, however, direct pool-based selection remains feasible and provides a useful control that does not introduce a separate learned matcher. Conversely, whole-pool prompting becomes impractical as the pool grows and may exacerbate long-context and position-bias effects. Batch-tournament limits the size of each prompt, but its number of LLM calls still grows with the pool size and may become prohibitively expensive.

\paragraph{Cost considerations.}

Inference costs also limit the use of larger proprietary models. Although such models might yield more convincing results, their API costs can quickly become prohibitive, particularly for LLM matching and batch-tournament methods. Nevertheless, these expenses should ultimately be weighed against the cost of physical experiments.

\section{Implementation details}
\label{app:implementation-details}

\subsection{Model information}
\label{app:model-settings}

\gemmamodel{} is a dense 31B-parameter model \citep{gemma4}. \qwenmid{} is also a dense model, with 27B parameters \citep{qwen35blog}. The remaining models use mixture-of-experts architectures: \deepseekmodel{} has 284B total parameters, of which 13B are activated per token \citep{deepseekv4}. \qwenmidmoe{} has 35B total and 3B activated parameters, while \qwenbig{} has 397B total and 17B activated parameters \citep{qwen35blog}.

All checkpoints were served locally with vLLM and the \qwenfamily{} models use the official FP8 checkpoints. In the main experiments, temperature is set to zero, which corresponds to greedy decoding, and thinking was disabled for every model.

\paragraph{\qwenbig{} proxies}

Due to large inference costs and computational requirements, we could not complete all 25 seeds for each of the whole-pool \qwenbig{} runs. We therefore use proxies of batch-tournament with $b=300$ indicated by \(^{*}\), which is close to the pool size of 312 of \steelsdataset{}, but far from the 921 of \feconidataset{}. Moreover, \(^{\dagger}\) denotes similar proxies of $b=32$ instead of $b=20$. For \electrostraindataset{}, we have the complete runs, so we do not need these proxies.  These proxies are used in Table~\ref{tab:main-results}, Table~\ref{tab:position-bias-metrics-full}, Figure~\ref{fig:position-bias-metrics}, and explain why some points are missing for \qwenbig{} in Figure~\ref{fig:batch-size}.

\subsection{Datasets}

\paragraph{Electrostrain.}
\label{app:electrostrain}

Each Electrostrain candidate is represented by eight composition variables and seven auxiliary composition-derived descriptors. The LLM policies receive all 15 features. For GP-EI, we evaluated both the full 15-dimensional representation and the eight composition variables alone, using otherwise identical preprocessing and GP settings. In preliminary experiments, the full-feature GP-EI performed similarly to random selection with a mean iterations to optimum of $37.32 \pm 22.30$, instead of $18.48 \pm 14.53$ for the composition-only representation, which remained comparable to the LLM policies. We therefore use the composition-only GP-EI in the primary comparisons.


\subsection{Policies}
\label{app:policies}

\paragraph{Gaussian process.}

In our implementation, the GP is refit from scratch at every loop iteration. Feature vectors are standardized coordinate-wise using the mean and standard deviation of the observed feature matrix, and the same transformation is applied to the unobserved pool. Targets are also standardized using the observed-label mean and standard deviation before computing $\mathrm{EI}$. The reported $\mathrm{EI}$ ranking is therefore computed in normalized target units. Standard deviations are clipped below at $10^{-8}$ to handle zero-variance coordinates, while posterior standard deviations used inside $\mathrm{EI}$ are clipped below at $10^{-9}$ before evaluating $z_t(x)$. We use scikit-learn's \texttt{GaussianProcessRegressor} with an isotropic \texttt{Matérn} kernel and length-scale bounds $(10^{-10},10^5)$; the Matérn length scale is fitted by marginal-likelihood optimization with the default L-BFGS-B optimizer and five optimizer restarts, and we do not add an explicit learned noise kernel.

\paragraph{Batch-tournament.}

At each active-learning iteration, the remaining unqueried candidates are randomly permuted and split into consecutive batches. Each full batch contains exactly $b$ candidates, while a final remainder batch contains fewer than $b$ when the pool size is not divisible by $b$. A singleton remainder is therefore treated as a one-candidate batch and advances unchanged. Every LLM call receives the same observed history. Within each batch, the candidates are displayed in a random order, and the model retains one winner by returning its displayed index. The winners are collected in batch order and are not globally reshuffled before being partitioned into the next round of batches, even though candidate order within each new LLM call is randomized again. This procedure continues until at most $b$ candidates remain, at which point a final LLM call selects the candidate to be evaluated by the oracle. Figure~\ref{fig:batch-tournament-schematic} illustrates the procedure without the various shuffles.
\input{tables/batch-tournament.tex}

\begin{figure}[t]
  \centering
  \includegraphics[width=\linewidth]{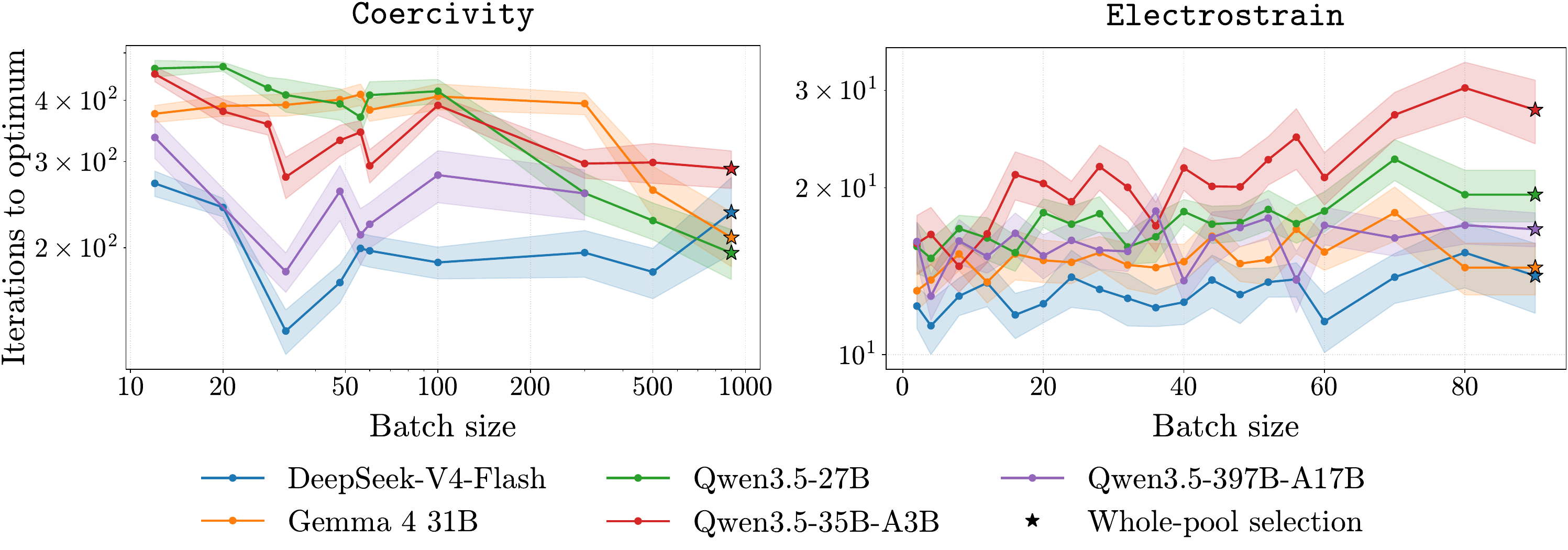}
  \caption{Batch-size sensitivity of tournament selection on \coercivitydataset{} and \electrostraindataset{}. Curves report the mean number of iterations to the optimum across seeds, with uncertainty shown shaded as the mean
    $\pm$ $\mathrm{std}/\sqrt{n_{\mathrm{seeds}}}$.}
  \label{fig:batch-size-coer}
\end{figure}

Figure~\ref{fig:batch-size-coer} presents the batch-size sweeps for \coercivitydataset{} and \electrostraindataset{}, whose trends are similar to those observed for the two tasks analyzed in the main text. Several \coercivitydataset{} configurations are omitted because their inference times were prohibitively long.

\subsection{Prompts.}
\label{app:prompts}

\input{prompts-steels}

\subsection{Position bias}
\label{app:position-bias}

\paragraph{Implementation.}
Because the available pool shrinks after every acquisition, raw indices are not directly comparable across iterations. For a selected candidate at one-indexed position $i$ in a displayed pool of size $K$, we define its normalized position as
\begin{equation}
  z = \frac{i-1}{K-1} \in [0,1].
\end{equation}

Very small remaining pools provide fewer opportunities for position bias to appear. We therefore retain only selections made before more than 100 candidates have been removed when considering whole-pool selection, i.e., while $K \geq N-100$ for an initial pool of size $N$. On batch-tournament, we are only considering requests where the LLM had to choose in a batch whose size is strictly equal to the batch size.

Let $q_b$ be the empirical probability mass of the retained normalized positions in bin $b$, using a fixed set of $n_{\mathrm{bins}}$ equal-width bins on $[0,1]$. We quantify departure from this uniform distribution using
\begin{equation}
  D_{\mathrm{KL}}(q\,\|\,u)
  =
  \sum_{b=1}^{n_{\mathrm{bins}}}
  q_b \log\frac{q_b}{u_b}
  =
  \sum_{b=1}^{n_{\mathrm{bins}}}
  q_b \log\!\left(n_{\mathrm{bins}}q_b\right).
\end{equation}
with the convention $0\log 0=0$.

We use 150 bins for the calculations reported in the table and 80 bins for the displayed plots. Because of that, we have to be careful when handling the results: we cannot directly compare batch-tournament $D_{\mathrm{KL}}(q\,\|\,u)$ values to those of whole-pool selection.

\paragraph{Additional results.}
Table~\ref{tab:position-bias-metrics-full} extends the comparison to every available model-task pair by reporting $D_{\mathrm{KL}}(q\,\|\,u)$ for every model and task. \deepseekmodel{} leads almost consistently, while \gemmamodel{} remains the most consistent by always achieving reasonable $\mathrm{KL}$-divergence values. Within the \qwenfamily{} model family, the global behavior remains close to that of \gemmamodel{}. Regardless, trends are similar to those highlighted in the main results.

\input{tables/position-bias}

\section{Additional implementations}
\label{app:additional-implementations}

\subsection{Additional policies}
\label{sec:generation-and-matching}

We have tested three other LLM policies that did not yield convincing results in our configuration. Thus, we decided to not include them in the paper.

\paragraph{GP+LLM.}

Following the general idea of hybrid LLM Bayesian optimization methods, we have tested a Bayesian optimization implementation that uses both LLM and GP-EI. At each iteration, we considered the top-$p$ best $\mathrm{EI}$ samples (with $p$ varying from 2 up to 16), and asked the LLM to choose the most promising one. It consistently gave similar results to a random choice among the points given by the GP-EI. This remark goes along those of section~\ref{sec:gpdiagnostics}, as well as further experiments of Section~\ref{par:gpsub}, which suggest that LLM selections do not follow the same exploration-exploitation pattern as GP-EI.

\paragraph{Swiss-tournament.}

We have also tried a Swiss-style tournament, where the winners of each round are paired against other winners in the next round, but it did not significantly improve performance over the batch-tournament approach, while causing longer inference times. We therefore focused on the batch-tournament and whole-pool protocols in our main experiments.

\paragraph{Generation-and-matching.}
Inspired by \citet{wang2026trainingfree}, we include a generation-and-matching protocol. When presented the observed history, the LLM generates a new candidate composition, before asking a model (which might not be the same as the generating LLM) to match the generated candidate to one of the unobserved pool candidates.

Unlike \citet{wang2026trainingfree}, who use the specialized Cohere Rerank-v3.5 model and its native relevance scores \citep{cohere2024rerank}, we perform matching with the same underlying model we evaluate, using normalized first-token log probabilities for \texttt{yes} and \texttt{no}.

For each generated proposal $q$, a reranking model compares it with every unobserved candidate $x$ and is prompted to answer \texttt{yes} or \texttt{no} according to the similarity of their elements and fractions. We request one output token and define the matching score from its log probabilities as
\[
  s(q,x)=\frac{\exp(\operatorname*{logprob}(\texttt{yes}))}
  {\exp(\operatorname*{logprob}(\texttt{yes}))+\exp(\operatorname*{logprob}(\texttt{no}))}.
\]

Missing \texttt{yes} or \texttt{no} tokens in the returned top-20 are assigned a log probability of $-100$, and the candidate maximizing $s(q,x)$ is selected. All pairs are scored concurrently through the vLLM OpenAI-compatible API, proposals are truncated to 500 characters and reasoning is disabled.

Table~\ref{tab:generation-matching-results} reports the results on \steelsdataset{}. With \gemmamodel{}, generation-and-matching performs similarly to whole-pool selection on average, although with slightly greater variability. With \qwenmid{}, it requires approximately 2.6 times the mean number of iterations of whole-pool selection and also exhibits a substantially larger standard deviation. Interestingly, we do not observe the same behavior as that of \citet{wang2026trainingfree}. Under the settings evaluated here, generation-and-matching performs worse on average than both GP-EI and whole-pool selection (see Appendix~\ref{sec:limitations}). We therefore focus on whole-pool selection in the main analysis.

\input{tables/generation-matching-results}

\subsection{Potential improvements}

\paragraph{History management.}
To limit context growth, we tested an LLM-based history-management strategy that retains at most \(s\) observations by selecting which observations to keep at each iteration. This strategy performed comparably to the standard batch-tournament protocol on \coercivitydataset{} for \(s=64\) and \(s=128\). Other selection rules, including geometric approaches, may also be worth exploring. However, we did not obtain sufficiently long runs to evaluate the method reliably beyond approximately 300 iterations, so we excluded it from the main results.

\paragraph{Chain-of-Thought reasoning.}

Chain-of-Thought prompting can improve LLM reasoning \citep{wei2022chain}, but its computational cost prevented us from completing a batch-tournament evaluation. We therefore exclude it from the reported comparisons. Evaluating whether explicit reasoning improves generation-and-matching in larger candidate pools remains an avenue for future work.

\section{Additional diagnostics}

\paragraph{UCB coefficient fit.}
\label{sec:ucb-fit}

To further explore the behavior of the LLM, we introduce the Upper Confidence Bound ($\operatorname{UCB}$) acquisition function, which selects the candidate maximizing
\[
  \operatorname{UCB}_{\beta}(x)=\mu(x)+\sqrt{\beta}\,\sigma(x),
\]
where $\mu(x)$ is the predicted performance and $\sigma(x)$ the predictive uncertainty. The parameter $\beta$ controls the exploration-exploitation trade-off: small values favor exploitation, while large values favor exploration.

To quantify the exploration-exploitation trade-off of each method, we associate each decision with an equivalent $\operatorname{UCB}$ coefficient $\beta$: at each iteration, we fit a GP from the observations available before the decision, using the same settings as our previous GP-EI. We then estimate the value $\beta \geq 0$ for which the selected point is as close as possible to maximizing $\operatorname{UCB}$:
\[
  \widehat{\beta}_t
  =
  \min\!\left(
  \underset{\beta \geq 0}{\operatorname{argmin}}
  \left\{
  \max_{x\in\{1,\ldots,N\} \setminus \mathcal{I}_t}
  \operatorname{UCB}_{\beta}(x_i)
  -
  \operatorname{UCB}_{\beta}(x_{i_t^{\mathrm{LLM}}})
  \right\}
  \right).
\]

\begin{figure}[t]
  \centering
  \includegraphics[width=\linewidth]{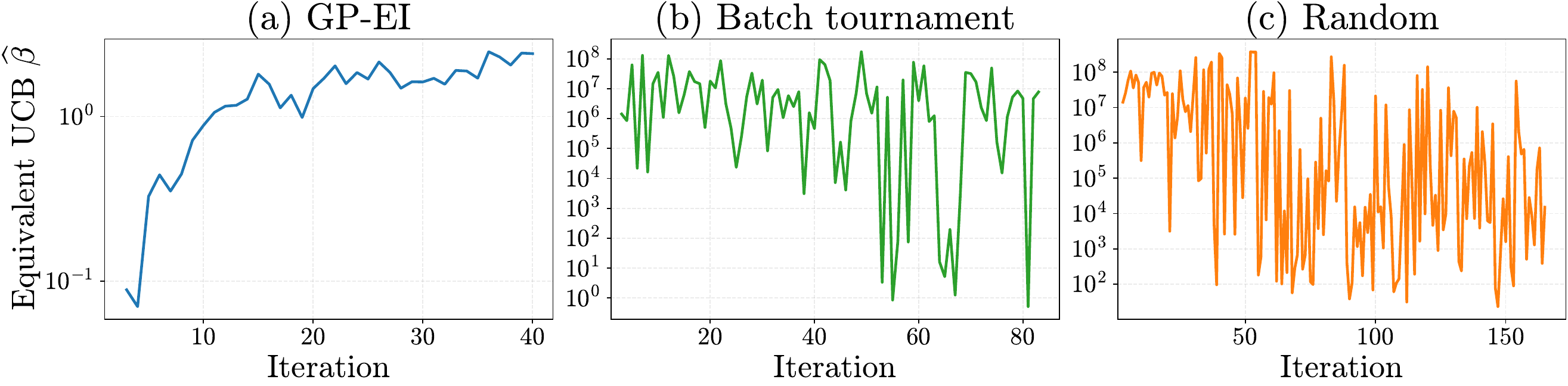}
  \caption{Comparison of evolution of fitted $\widehat{\beta}_t$ on \steelsdataset{} between \gemmamodel, Random and GP-EI. Results are averaged across seeds for each iteration, when at least half of the seeds have not converged yet.}
  \label{fig:fit-ucb}
\end{figure}

Figure~\ref{fig:fit-ucb} examines this behavior across seeds through the fitted $\operatorname{UCB}$ coefficient $\widehat{\beta}_t$. GP-EI exhibits a comparatively stable UCB-equivalent profile, while the coefficients fitted to \gemmamodel{} fluctuate more strongly and are closer to those obtained under random selection. Thus, although LLM selections are informed enough to outperform random selection in the main evaluation, their decisions are not consistently approximated by a fixed or smoothly evolving GP-UCB-like policy.

\paragraph{GP Substitution.}
\label{par:gpsub}

Instead of fitting a $\operatorname{UCB}$ parameter, we have also tried observing the rank in terms of $\mathrm{EI}$ of the LLM-selected candidate after fitting a GP. Even though this method provided consistent results with those obtained previously, it was too arbitrary, which is why we decided to not include it in the main results.

\section{AI usage} We used AI assistants during the development and writing process, including for code prototyping, debugging, experiment-management scripts, and language polishing. All experimental results, analyses, and paper claims were checked by the authors.

\end{document}

%% file: tables/datasets.tex
\begin{table}[t]
  \centering
  \caption{Retrospective optimization tasks.}
  \label{tab:datasets}
  \begin{tabular}{lrrl}
    \toprule
    Task                  & Candidates & Features & Target                    \\
    \midrule
    \kerrdataset          & 921        & 3        & Kerr rotation (mrad) \\
    \coercivitydataset    & 921        & 3        & coercivity (mT)      \\
    \steelsdataset        & 312        & 14       & yield strength (MPa) \\
    \electrostraindataset & 81         & 15       & electrostrain (\%)       \\
    \bottomrule
  \end{tabular}
\end{table}

%% file: tables/main-results.tex
\begin{table}[t]
  \centering
  \caption{Iterations to the global optimum over 25 initializations (mean $\pm$ standard deviation), except for random selection which is initialized over 1000 seeds. Lower is better. Baselines are shown once because they do not depend on an LLM presentation protocol. }
  \label{tab:main-results}
  \small
  \setlength{\tabcolsep}{3.5pt}
  \renewcommand{\arraystretch}{1.08}
  \begin{tabular*}{\textwidth}{@{\extracolsep{\fill}}lcccc@{}}
    \toprule
    Model
    & \textbf{\kerrdataset{}}
    & \textbf{\coercivitydataset{}}
    & \textbf{\steelsdataset{}}
    & \textbf{\electrostraindataset{}} \\
    \midrule
    \textsc{Random}
    & \meanstd{462.48}{262.24}
    & \meanstd{448.09}{269.63}
    & \meanstd{159.31}{91.28}
    & \meanstd{39.89}{23.00} \\
    \textsc{GP-EI}
    & \meanstd{8.40}{2.78}
    & \meanstd{91.80}{62.77}
    & \meanstd{42.72}{27.24}
    & \meanstd{18.48}{14.53} \\
    \midrule
    \multicolumn{5}{@{}l}{\emph{Whole-pool selection}} \\
    \gemmamodelshort
    & \meanstd{31.28}{17.24}
    & \meanstd{209.88}{134.05}
    & \meanstd{44.16}{23.36}
    & \meanstd{14.36}{7.71} \\
    \qwenmidshort
    & \meanstd{27.08}{13.59}
    & \meanstd{195.20}{114.39}
    & \meanstd{48.24}{23.43}
    & \meanstd{19.44}{10.40} \\
    \qwenmidmoeshort
    & \meanstd{45.72}{20.28}
    & \meanstd{289.88}{127.44}
    & \meanstd{50.36}{24.89}
    & \meanstd{27.64}{18.15} \\
    \qwenbigshort
    & \meanstd{23.48}{10.40}$^\ast$
    & \meanstd{258.24}{151.68}$^\ast$
    & \meanstd{53.32}{32.90}$^\ast$
    & \meanstd{16.84}{5.93} \\
    \deepseekmodelshort
    & \meanstd{110.64}{85.31}
    & \meanstd{236.28}{210.72}
    & \meanstd{87.60}{52.10}
    & \meanstd{13.88}{9.98} \\
    \midrule
    \multicolumn{5}{@{}l}{\emph{Batch-tournament ($b=20$)}} \\
    \gemmamodelshort
    & \meanstd{13.12}{5.12}
    & \meanstd{389.64}{92.00}
    & \meanstd{78.08}{49.83}
    & \meanstd{14.80}{6.50} \\
    \qwenmidshort
    & \meanstd{14.28}{7.33}
    & \meanstd{468.64}{50.68}
    & \meanstd{64.52}{32.94}
    & \meanstd{18.04}{5.33} \\
    \qwenmidmoeshort
    & \meanstd{22.00}{8.73}
    & \meanstd{379.80}{110.27}
    & \meanstd{52.68}{20.12}
    & \meanstd{20.36}{9.89} \\
    \qwenbigshort
    & \meanstd{14.48}{7.54}
    & \meanstd{178.80}{82.87}$^\dagger$
    & \meanstd{81.96}{41.10}
    & \meanstd{15.08}{6.92} \\
    \deepseekmodelshort
    & \meanstd{16.88}{5.52}
    & \meanstd{241.84}{52.10}
    & \meanstd{32.88}{31.71}
    & \meanstd{12.36}{4.71} \\
    \bottomrule
  \end{tabular*}
\end{table}

%% file: tables/position-bias-steels.tex
\centering
\small
\setlength{\tabcolsep}{4pt}
\renewcommand{\arraystretch}{1.25}
\begin{tabular}{lc}
  \toprule
  \textbf{Models} & \textbf{$D_{\mathrm{KL}}(q\,\|\,u)$} \\
  \midrule
  \gemmamodelshort      & 1.02 \\
  \deepseekmodelshort   & 1.47 \\
  \qwenmidshort         & 0.852 \\
  \qwenmidmoeshort      & 0.747 \\
  \qwenbigshort         & 1.36$^\ast$ \\
  \bottomrule
\end{tabular}

%% file: tables/batch-tournament.tex
\begin{figure}[!h]
  \centering
  \begin{tikzpicture}[
    font=\scriptsize,
    cand/.style={
        draw,
        rounded corners=1pt,
        minimum width=0.62cm,
        minimum height=0.40cm,
        inner sep=1pt,
        fill=gray!8
      },
    win/.style={
        cand,
        fill=blue!12,
        draw=blue!65!black,
        very thick
      },
    batch/.style={
        draw=gray!55,
        rounded corners=2pt,
        inner sep=3pt,
        dashed
      },
    prompt/.style={
        draw,
        rounded corners=2pt,
        minimum width=1.65cm,
        minimum height=0.43cm,
        inner sep=2pt,
        fill=gray!12
      },
    arrow/.style={-{Latex[length=2mm]}, thick, draw=gray!70}
    ]

    \node[anchor=east] at (-0.55,0) {Pool};
    \foreach \i/\x in {
        1/0,2/0.72,3/1.44,4/2.16,
        5/3.20,6/3.92,7/4.64,8/5.36,
        9/6.40,10/7.12,11/7.84,12/8.56,
        13/9.60,14/10.32,15/11.04,16/11.76
      } {
        \node[cand] (x\i) at (\x,0) {$x_{\i}$};
      }

    \node[batch, fit=(x1)(x2)(x3)(x4)] (b1) {};
    \node[batch, fit=(x5)(x6)(x7)(x8)] (b2) {};
    \node[batch, fit=(x9)(x10)(x11)(x12)] (b3) {};
    \node[batch, fit=(x13)(x14)(x15)(x16)] (b4) {};

    \foreach \j/\center in {1/1.08,2/4.28,3/7.48,4/10.68} {
        \node[prompt] (p\j) at (\center,-0.95) {LLM pick};
        \draw[arrow] (\center,-0.23) -- (p\j.north);
      }

    \node[anchor=east] at (-0.55,-1.85) {Winners};
    \node[win] (w1) at (1.08,-1.85) {$w_1$};
    \node[win] (w2) at (4.28,-1.85) {$w_2$};
    \node[win] (w3) at (7.48,-1.85) {$w_3$};
    \node[win] (w4) at (10.68,-1.85) {$w_4$};
    \foreach \j in {1,...,4} {
        \draw[arrow] (p\j.south) -- (w\j.north);
      }

    \node[batch, fit=(w1)(w2)(w3)(w4)] (finalbatch) {};
    \node[prompt] (finalprompt) at (5.88,-2.85) {final LLM pick};
    \draw[arrow] (finalbatch.south) -- (finalprompt.north);

    \node[win, minimum width=1.05cm] (winner) at (5.88,-3.75) {$x_t$};
    \node[prompt, right=0.45cm of winner, minimum width=1.40cm] (oracle) {oracle label};
    \draw[arrow] (finalprompt.south) -- (winner.north);
    \draw[arrow] (winner.east) -- (oracle.west);

    \node[prompt, below=0.35cm of oracle, minimum width=2.35cm, align=center]
    (update) {update $\mathcal{D}_{t+1}$};
    \draw[arrow] (oracle.south) -- (update.north);
  \end{tikzpicture}
  \caption{Batch-tournament candidate selection for an example with $N=16$ unobserved candidates and batch size $b=4$.
  }
  \label{fig:batch-tournament-schematic}
\end{figure}

%% file: prompts-steels.tex
We report below the prompt templates used for \steelsdataset{} under the
materials-aware and anonymized conditions for batch-tournament selection.

\Needspace{14\baselineskip}
\subsubsection{Materials-aware.}

\promptheading{System prompt.}
\begin{promptblock}
  You are a materials scientist optimizing steel alloy compositions for maximum yield strength.

  You will be shown previously tested compositions and their yield strengths (in MPa), then a numbered list of candidate compositions that have not been tested yet.
  Decide which candidate is most promising to test next to find high yield strength.

  **CRITICAL: Respond with ONLY the number of your chosen candidate. Nothing else.**
  Do NOT include any text, reasoning, or additional commentary.
  Your entire response must be a single number: 1, 2, 3, 4, 5, 6, 7, 8, etc.
\end{promptblock}

\promptheading{User prompt.}
\begin{promptblock}
  Tested compositions and their yield strengths:

  {
  {{ loop.index }}. {{ comp }} -> {{ "
  {
  Candidate compositions:

  {
  {{ loop.index }}. {{ comp }}
  {
  Which candidate should be tested next?
\end{promptblock}

\subsubsection{Anonymized}

\promptheading{System prompt.}
\begin{promptblock}
  You are an optimization algorithm searching for the global maximum of an unknown black-box function.

  You will be shown previously tested input vectors and their numeric scores, then a batch of candidate input vectors that have not been tested yet.
  Decide which candidate is most promising to test next to maximize the score.

  **CRITICAL: Respond with ONLY the number of your chosen candidate. Nothing else.**
  Do NOT include any text, reasoning, or additional commentary.
  Your entire response must be a single number: 1, 2, 3, 4, 5, etc.
\end{promptblock}

\promptheading{User prompt.}
\begin{promptblock}
  Tested input vectors and their scores:

  {
  {{ loop.index }}. {{ vec }} -> {{ "
  {
  Candidate input vectors:

  {
  {{ loop.index }}. {{ vec }}
  {
  From the candidates above, choose the one most likely to produce the highest score. Answer with exactly the number of the chosen candidate.
\end{promptblock}

%% file: tables/position-bias.tex
\begin{table*}[t]
  \centering
  \caption{Position-bias metrics for all model-task combinations. We report $D_{\mathrm{KL}}$ for whole-pool and batch-tournament selection with $b=20$, except for the proxy configurations marked with an asterisk. Missing measurements are denoted by '-'.}

  \label{tab:position-bias-metrics-full}
  \small
  \setlength{\tabcolsep}{3pt}
  \renewcommand{\arraystretch}{1.05}
  \begin{tabular*}{\textwidth}{@{\extracolsep{\fill}}l*{8}{c}@{}}
    \toprule
     & \multicolumn{2}{c}{\textbf{\electrostraindataset}}
     & \multicolumn{2}{c}{\textbf{\steelsdataset}}
     & \multicolumn{2}{c}{\textbf{\coercivitydataset}}
     & \multicolumn{2}{c}{\textbf{\kerrdataset}} \\
    \cmidrule(lr){2-3}\cmidrule(lr){4-5}
    \cmidrule(lr){6-7}\cmidrule(lr){8-9}
    \textbf{Models}
    & \textbf{Whole} & \textbf{Batch}
    & \textbf{Whole} & \textbf{Batch}
    & \textbf{Whole} & \textbf{Batch}
    & \textbf{Whole} & \textbf{Batch} \\
    \midrule
    \gemmamodelshort
    & 0.998 & 2.27
    & 1.02  & 2.11
    & 1.60  & 2.08
    & 2.21 & 2.02 \\
    \addlinespace
    \deepseekmodelshort
    & 1.36 & 2.29
    & 1.47 & 2.13
    & 2.48 & 2.35
    & 2.60 & 2.15 \\
    \addlinespace
    \qwenmidshort
    & 0.929 & 2.10
    & 0.852 & 2.04
    & 2.18 & 2.06
    & 2.17 & 2.04 \\
    \addlinespace
    \qwenmidmoeshort
    & 1.26  & 2.20
    & 0.747 & 2.12
    & 1.67  & 2.24
    & 2.77 & 2.08 \\
    \addlinespace
    \qwenbigshort
    & 1.04 & 2.20
    & 1.36$^\ast$ & 2.08
    & -- & 1.84$^\dagger$
    & -- & 2.06 \\
    \bottomrule
  \end{tabular*}
\end{table*}

%% file: tables/generation-matching-results.tex
\begin{table}[t]
  \centering
  \caption{Generation-and-matching performance on \steelsdataset{} compared with random, GP-EI, and whole-pool selection. We report iterations to the global optimum over 25 initializations (mean $\pm$ standard deviation), except for random selection which is initialized over 1000 seeds.}
  \label{tab:generation-matching-results}
  \small
  \setlength{\tabcolsep}{6pt}
  \renewcommand{\arraystretch}{1.08}
  \begin{tabular*}{\textwidth}{@{\extracolsep{\fill}}lcccc@{}}
    \toprule
    Model
    & \textbf{Random}
    & \textbf{GP-EI}
    & \textbf{Whole-pool}
    & \textbf{Generation-and-matching} \\
    \midrule
    \qwenmidshort
    & \multirow{2}{*}{\meanstd{159.31}{91.28}}
    & \multirow{2}{*}{\meanstd{42.72}{27.24}}
    & \meanstd{48.24}{23.43}
    & \meanstd{123.60}{63.60}
    \\
    \gemmamodelshort
    &
    &
    & \meanstd{44.16}{23.36}
    & \meanstd{48.80}{26.14}
    \\
    \bottomrule
  \end{tabular*}
\end{table}